\documentclass[runningheads]{llncs}
\usepackage[T1]{fontenc}
\usepackage{graphicx,verbatim}
\usepackage{hyperref}
\usepackage{color}

\usepackage{caption}
\usepackage{subcaption}
\usepackage{amsmath}
\usepackage{amssymb}
\usepackage{tabularx}
\usepackage{multirow}
\usepackage{booktabs}
\usepackage{graphicx}
\usepackage{array}

\definecolor{gray}{rgb}{0.4,0.4,0.4}
\definecolor{brown}{rgb}{0.59,0.29,0}
\definecolor{green_good}{rgb}{0,0.69,0.31}
\definecolor{red_bad}{rgb}{0.75,0,0}

\begin{document}
%
\title{Toward Using Machine Learning as a Shape Quality Metric for Liver Point Cloud Generation}
\titlerunning{Liver Point Cloud Shape Classification}
%



\author{Khoa Tuan Nguyen\inst{1,2} \and
Gaeun Oh\inst{2} \and
Ho-min Park\inst{1,2} \and
Francesca Tozzi\inst{4,5} \and
Wouter Willaert\inst{4,6} \and
Joris Vankerschaver\inst{2,3} \and
Niki Rashidian\inst{4,7} \and
Wesley De Neve\inst{1,2}}

\authorrunning{Khoa et al.}

\institute{
IDLab, ELIS, Ghent University, Ghent, Belgium \and
Center for Biosystems and Biotech Data Science, Ghent University Global Campus, Incheon, Korea \and
Department of Mathematics, Computer Science and Statistics, Ghent University, Ghent, Belgium \\
\email{\{khoatuan.nguyen,gaeun.oh,homin.park,\\joris.vankerschaver,wesley.deneve\}@ghent.ac.kr} \and 
Department of Human Structure and Repair, Ghent University, Ghent, Belgium \and
Department of Hepatobiliary Surgery, Centre Hospitalier de l'Université de Montréal, Canada \and
Department of GI Surgery, Ghent University Hospital, Ghent, Belgium \and
Department of HPB Surgery \& Liver Transplantation, Ghent University Hospital, Ghent, Belgium \\
\email{\{francesca.tozzi,wouter.willaert,nikdokht.rashidian\}@ugent.be}
}

\maketitle              

\setcounter{footnote}{0}

\begin{abstract}
While 3D medical shape generative models such as diffusion models have shown promise in synthesizing diverse and anatomically plausible structures, the absence of ground truth makes quality evaluation challenging.
Existing evaluation metrics commonly measure distributional distances between training and generated sets, while the medical field requires assessing quality at the individual level for each generated shape, which demands labor-intensive expert review.

In this paper, we investigate the use of classical machine learning (ML) methods and PointNet as an alternative, interpretable approach for assessing the quality of generated liver shapes.
We sample point clouds from the surfaces of the generated liver shapes, extract handcrafted geometric features, and train a group of supervised ML and PointNet models to classify liver shapes as good or bad.
These trained models are then used as proxy discriminators to assess the quality of synthetic liver shapes produced by generative models.

Our results show that ML-based shape classifiers provide not only interpretable feedback but also complementary insights compared to expert evaluation.
This suggests that ML classifiers can serve as lightweight, task-relevant quality metrics in 3D organ shape generation, supporting more transparent and clinically aligned evaluation protocols in medical shape modeling.


\end{abstract}
%
%
%
%
%
%
%
\section{Introduction}

\begin{figure}[t]
    \centering
    \includegraphics[width=1\linewidth]{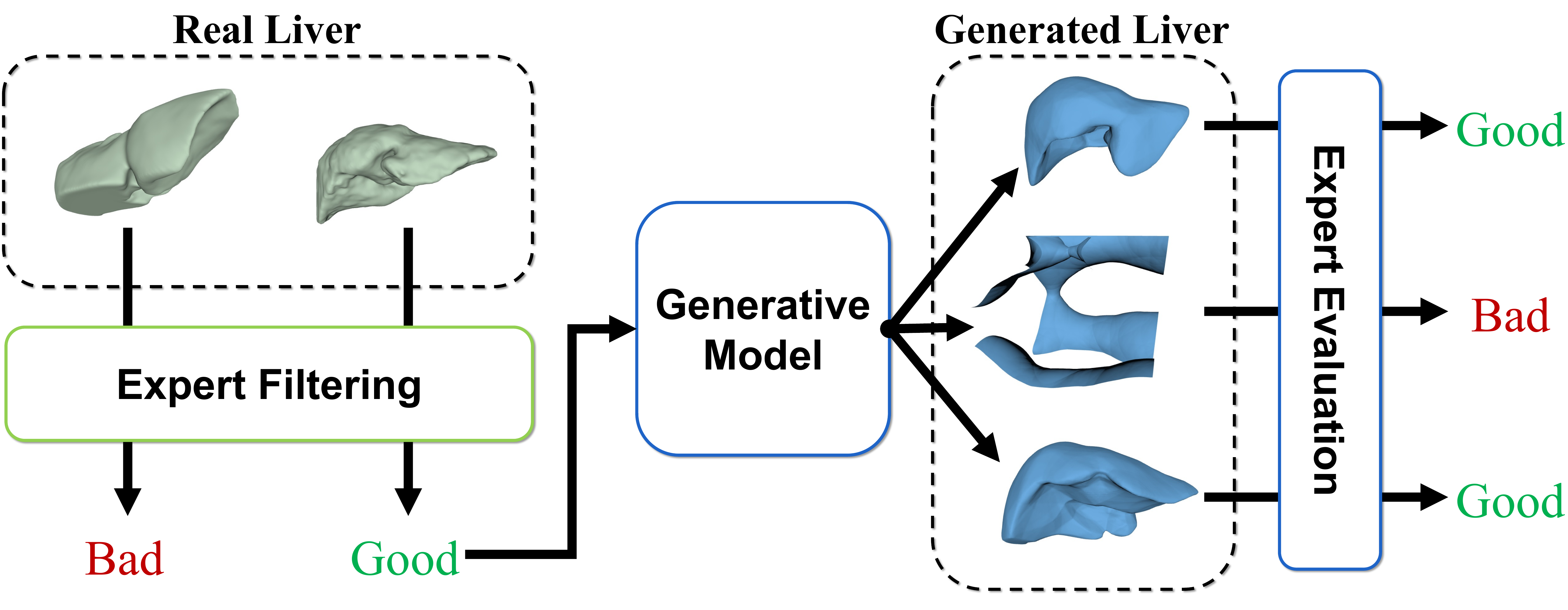}
    \caption{
    Workflow implemented by our prior work~\cite{nguyen2025boosting}.
    Given raw liver objects, we filtered and used the good liver objects to train a generative model to synthesize livers.
    Filtering good liver objects (``Expert Filtering'') and evaluating all generated livers (``Expert Evaluation'') requires substantial effort from our surgical team.
    In this paper, we apply ML methods during the ``Expert Evaluation'' step to reduce manual labor, support the experts, and accelerate our workflow.
    }
    \label{fig:previous_work}
\end{figure}

With the open availability of large-scale datasets such as Laion-5B~\cite{schuhmann2022laion} and Objaverse-XL~\cite{deitke2024objaverse}, along with the success of generative models such as diffusion models, recent advancements have been made in 3D generation~\cite{lee2024text,po2024state} and 3D reconstruction~\cite{yunus2024recent}.
To apply these advancements to the medical field, a large amount of data is also required, but due to patient privacy, access to datasets is limited.
However, several open 3D medical shape datasets have been released, including SARAMIS~\cite{montana2023saramis}, MedShapeNet~\cite{li2025medshapenet}, and AbdomenAtlas~\cite{li2024abdomenatlas}, which has enabled many works in 3D medical shape generation~\cite{jayakumar2023sadir,li2023anatomy,li2025medshapenet}.

Our prior research~\cite{nguyen2025boosting} discovered substantial quality inconsistencies in publicly available 3D liver shape datasets, with only 48.14\% of the liver objects in TotalSegmentator being usable due to incomplete shapes, artifacts, and non-liver structures. 
To address these limitations, we collaborated with medical experts at Ghent University Hospital to establish a quality-controlled liver dataset with clinical annotations.
Fig.~\ref{fig:previous_work} illustrates our prior workflow for obtaining well generated liver objects from raw real liver objects\footnote{We use \textbf{`object'} to distinguish the 3D model from the term \textbf{`model'} in deep learning.}.
For simplicity, we use the annotation `Good' to represent good liver objects, and `Bad' to refer to objects that are not livers, are incomplete in shape, or contain artifacts, in both real and generated liver objects.
We identified two bottlenecks in this workflow: ``Expert Filtering'' and ``Expert Evaluation'', with the evaluation step being especially time-consuming due to the need for repeated assessments of newly generated liver objects.

In this paper, we explore the use of classical machine learning (ML) methods and PointNet~\cite{qi2017pointnet,qi2017pointnet++} as classifiers to mimic expert evaluation, aiming to reduce manual labor during the ``Expert Evaluation'' step and accelerate our workflow.

While point cloud quality assessment methods exist in computer vision~\cite{Liu2023Point,Shan_2024_CVPR,yang2022itpcqa} and medical shape evaluation frameworks rely on geometric metrics~\cite{han2023medgen3d,li2025medshapenet,Taha2015}, to the best of our knowledge, this is the first research effort to bridge these domains by applying ML classifiers trained on real medical point cloud data to evaluate synthetically generated medical shapes.

\section{Methods}

\begin{figure}[t]
    \centering
    \includegraphics[width=1\linewidth]{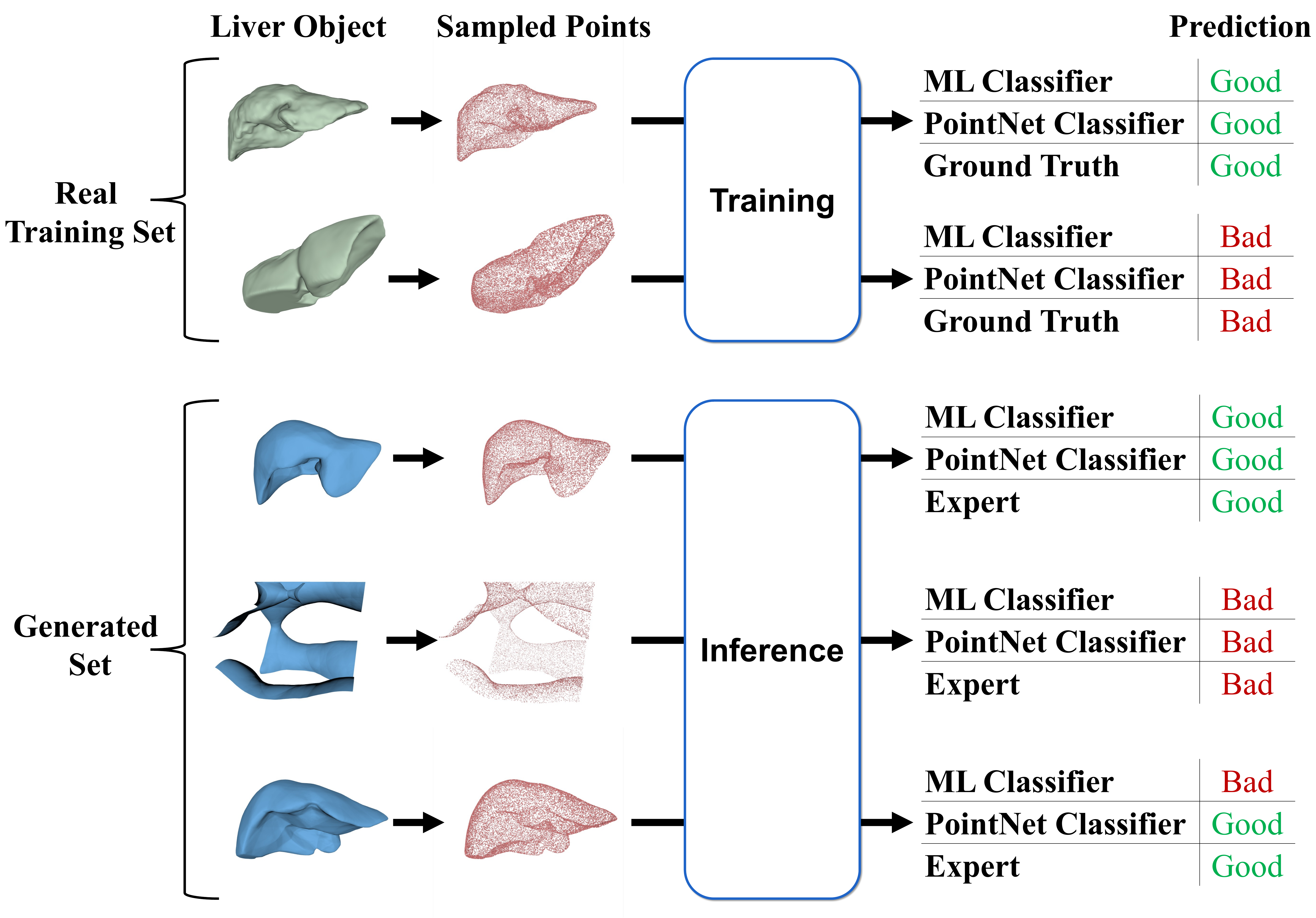}
    \caption{
    Overview of the proposed approach.
    In the first step, we train point cloud classifiers, including classical ML methods and PointNet~\cite{qi2017pointnet,qi2017pointnet++}.
    Next, in the inference step, we apply these trained classifiers to the liver objects generated by our prior work~\cite{nguyen2025boosting}.
    The predictions, including those from the ML classifiers and PointNet classifiers~\cite{qi2017pointnet,qi2017pointnet++}, are then compared with the expert assessments.
    Our method shows that by using only the `Good/Bad' labels from the real dataset, we can detect unseen error cases in generated liver objects, as demonstrated in the second row of the inference step.
    }
    \label{fig:overview}
\end{figure}

Our method consists of two steps: training the point cloud classifiers (Section~\ref{sec:pc_classification}) and performing inference on the generated liver objects (Section~\ref{sec:pc_evaluation}), as shown in Fig.~\ref{fig:overview}.

\subsection{Point Cloud Classification}
\label{sec:pc_classification}

\begin{figure}[tb]
    \centering
    \includegraphics[width=0.8\textwidth]{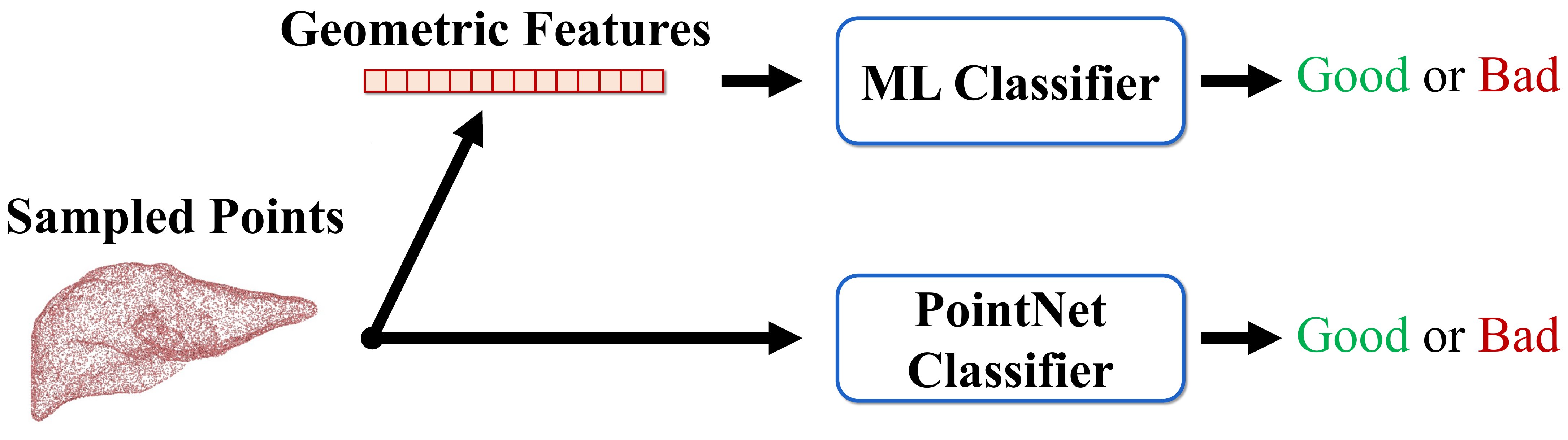}
    \caption{
    The training pipeline of each point cloud classifier takes as input a liver point cloud and outputs a `Good/Bad' prediction.
    For the ML classifiers, we extract 14-dimensional geometric features from the liver point cloud to reduce computational complexity.
    On the other hand, the PointNet-based classifiers operate directly on all 20,000-points.
    }
    \label{fig:pc_classifier}
\end{figure}


\begin{table}[t]
    \centering
    \caption{Geometric features extracted from each liver point cloud}
    \label{tab:extracted_features}
    \begin{tabularx}{\linewidth}{p{4cm}X}
        \toprule
        \textbf{Feature Name} & \textbf{Description} \\
        \midrule
        min\_x, min\_y, min\_z     & Minimum coordinates along x, y, and z axes \\
        max\_x, max\_y, max\_z     & Maximum coordinates along x, y, and z axes \\
        mean\_x, mean\_y, mean\_z & Mean coordinates along x, y, and z axes \\
        std\_x, std\_y, std\_z  & Standard deviation along x, y, and z axes \\
        mean\_radius & Mean distance from the origin ($r = \sqrt{x^2 + y^2 + z^2}$) \\
        std\_radius  & Standard deviation of distance from the origin \\
        \bottomrule
    \end{tabularx}
\end{table}

\textbf{Data Preparation and Feature Extraction.}
In our prior work~\cite{nguyen2025boosting}, hepatobiliary surgeons from Ghent University Hospital manually reviewed and categorized 939 liver objects exported from TotalSegmentator CT scans~\cite{wasserthal2023totalsegmentator} into five categories: `Usable' (48.14\%), `No full shape' (44.41\%), `Requires editing' (4.15\%), `Not usable' (2.98\%), and `Not sure' (0.32\%).
To mimic expert evaluation, we map `Good' to represent 452 `Usable' liver objects, and `Bad' to refer to 487 liver objects that are incomplete in shape or contain artifacts, corresponding to the four categories: `No full shape', `Requires editing', `Not usable', and `Not sure'.
This makes the dataset more balanced and reduces the task to a simpler binary classification problem using `Good/Bad' labels as ground truth.

As illustrated in Figure~\ref{fig:pc_classifier}, we randomly sample 20,000 points on the surface of each liver object to create the liver point cloud training dataset. 
These sampled point clouds serve as input for two different classification approaches: (1) classical ML classifiers that operate on 14-dimensional geometric feature vectors extracted from the point clouds, and (2) PointNet-based methods that directly process the raw 20,000-points without feature extraction. 

To reduce computational complexity from processing the full 20,000 points, the ML classifiers operate on compact 14-dimensional geometric feature vectors ($\in \mathbb{R}^{14}$) extracted from each liver point cloud, as detailed in Table~\ref{tab:extracted_features}. 
These 14 basic geometric features capture fundamental spatial properties of liver point clouds, including bounding box dimensions, centroid location, and overall size characteristics. While these simple statistical measures may not directly detect complex anatomical issues, they can identify gross shape abnormalities such as extreme size variations, unusual positioning, or severely incomplete objects that deviate substantially from typical liver geometry.

\textbf{Machine Learning (ML) Classifier.}
We utilize several classical ML classifiers implemented in scikit-learn~\cite{scikit-learn} to perform binary classification of liver point clouds.
The ML models include: Support Vector Machine (SVM)~\cite{cortes1995support}, Decision Tree~\cite{breiman1984cart}, AdaBoost~\cite{freund1997decision}, Random Forest~\cite{breiman2001random}, Extra Trees~\cite{geurts2006extremely}, Gradient Boosting~\cite{friedman2001greedy}, Multi-Layer Perceptron (MLP)~\cite{hinton1989connectionist}, k-NN classifier (KNN)~\cite{cover1967nearest}, Logistic Regression~\cite{cox1958regression}, and Linear Discriminant Analysis (LDA)~\cite{fisher1936use}.

\textbf{PointNet Methods.}
To compare with the ML classifiers, we also apply PointNet~\cite{qi2017pointnet} and PointNet++~\cite{qi2017pointnet++} as alternative classifiers. 
Unlike the ML classifiers that use extracted geometric features, these deep learning methods directly process the raw 20,000-points without feature extraction. 
The implementation is taken from~\cite{Pytorch_Pointnet_Pointnet2}.

\subsection{Evaluation of Generated Livers}
\label{sec:pc_evaluation}
After training the point cloud classifiers, we apply them to the livers produced by our generative model in our prior work~\cite{nguyen2025boosting}, with the aim of evaluating and comparing their predictions with the assessments made by the experts.
We want to emphasize that the classifiers are trained on the `Good/Bad' labels from real liver objects, while during the inference step, the generated set may contain unseen liver objects.
This allows us to test the robustness of each classifier.

\section{Experimental Results}
\textbf{Experimental Setup.}
Recalling Fig.~\ref{fig:previous_work} and the data preparation described in Section~\ref{sec:pc_classification}, we split the dataset into non-overlapping subsets: training ($80\%$), validation ($5\%$), and testing ($15\%$).
All the ML classifiers are trained using the default settings in scikit-learn.
We train PointNet and PointNet++ on the raw clouds of 20,000 points for $300$ epochs with a batch size of $32$, using the Adam optimizer and a learning rate of $1 \times 10^{-4}$ on a single A6000 GPU.

\subsection{Quantitative Results}

\begin{table}[t]
    \centering
    \caption{
    We report the accuracy and F1 score on the test set, and also measure the percentage of `Good/Bad' predictions on the generated set to calculate Cohen’s $\kappa$ score.
    The best value and follow-up value are denoted in \textbf{bold} and \underline{underlined}, respectively.
    }
    \label{tab:testset_generatedset_result}
    \begin{tabularx}{\linewidth}{
        p{3.5cm}                                    
        >{\centering\arraybackslash}p{1.25cm}       
        >{\centering\arraybackslash}p{1.25cm}       
        *3{>{\centering\arraybackslash}X}           
    }
        \toprule
        \multirow{2}{*}{\textbf{Model}} & 
        \multicolumn{2}{c}{\textbf{Test set}} & 
        \multicolumn{3}{c}{\textbf{Generated set}} \\
        \cmidrule(lr){2-3} \cmidrule(lr){4-6}
        & \textbf{Acc}~$\uparrow$ & \textbf{F1}~$\uparrow$ & \textbf{Good}~(\%) & \textbf{Bad}~(\%) & \textbf{Cohen's $\kappa$}~$\uparrow$ \\
        \midrule
        SVM~\cite{cortes1995support} & 0.8951 & 0.8950 & 95.24 & 4.76 & \underline{0.49} \\
        Decision Tree~\cite{breiman1984cart} & 0.8042 & 0.8034 & 77.78 & 22.22 & 0.11 \\
        AdaBoost~\cite{freund1997decision} & 0.8182 & 0.8171 & 77.78 & 22.22 & 0.11 \\
        Random Forest~\cite{breiman2001random} & \textbf{0.9091} & \textbf{0.9091} & 93.65 & 6.35 & 0.38 \\
        Extra Trees~\cite{geurts2006extremely} & \textbf{0.9091} & \textbf{0.9091} & 95.24 & 4.76 & \underline{0.49} \\
        Gradient Boosting~\cite{friedman2001greedy} & 0.8811 & 0.8810 & 92.06 & 7.94 & 0.32 \\
        MLP~\cite{hinton1989connectionist} & \underline{0.9021} & \underline{0.9021} & 93.65 & 6.35 & 0.38 \\
        KNN~\cite{cover1967nearest} & 0.8951 & 0.8949 & 96.83 & 3.17 & -0.02 \\
        Logistic Regression~\cite{cox1958regression} & 0.8601 & 0.8601 & 98.41 & 1.59 & -0.02 \\
        LDA~\cite{fisher1936use} & 0.8741 & 0.8740 & 95.24 & 4.76 & \underline{0.49} \\
        \midrule
        \textbf{PointNet}~\cite{qi2017pointnet} & 0.8741 & 0.8741 & 100.00 & 0.00 & 0.00 \\
        \textbf{PointNet++}~\cite{qi2017pointnet++} & 0.8671 & 0.8669 & 98.41 & 1.59 & \textbf{1.00} \\
        \midrule
        \textbf{Expert}  &&& 98.41 & 1.59 &\\
        \bottomrule
    \end{tabularx}
\end{table}

We evaluated all models on two datasets: the test set of real liver point clouds with ground truth labels, and a generated set consisting of 63 liver objects for which we assess predicted shape quality using the trained classifiers.
Table~\ref{tab:testset_generatedset_result} summarizes the classification accuracy and F1 scores on the test set, along with the percentage of predicted `Good/Bad' samples and Cohen's $\kappa$~\cite{cohen1960coefficient} agreement with expert labels on the generated set.

On the test set, the best-performing models in terms of accuracy and F1 score are Random Forest, Extra Trees, and MLP, all achieving over $90\%$ in both metrics. 
In contrast, Decision Tree and AdaBoost show noticeably lower performance, both with an accuracy of approximately $80\%$.
Compared to the ML classifiers, PointNet and PointNet++ achieved slightly lower performance on the test set, with accuracies of $87.4\%$ and $86.7\%$, respectively, which are below the best classical models.

For the generated set, most classifiers predict a high proportion of samples as `Good', ranging from $92.06\%$ (Gradient Boosting) to $98.41\%$ (Logistic Regression). 
However, the reliability of these predictions varies substantially as reflected by Cohen’s $\kappa$ score, which measures agreement with expert labels. 
Notably, SVM, Extra Trees, and LDA achieve the highest agreement scores (0.49), indicating moderate agreement with the expert.
Interestingly, KNN and Logistic Regression, while predicting a high proportion of `Good' shapes, show near-zero or even negative Cohen's $\kappa$, suggesting overconfident but poorly aligned predictions. 
In contrast, PointNet++ achieves perfect agreement ($\kappa = 1.00$), despite having a slightly lower test set accuracy ($86.7\%$), indicating strong alignment with the expert assessments.

\subsection{Qualitative Analysis}


\begin{table}[ht!]
    \centering
    \caption{
    Qualitative Results.
    Comparison of model predictions with expert evaluation.
    We illustrate two rendered views of each generated liver (zoom in on the digital version of the paper for a clearer view). 
    (a) and (b) show cases where all models agree with the expert, predicting ‘Good’ for a good liver shape and ‘Bad’ for a failed generated liver, indicating the classifier's ability to distinguish quality.
    In contrast, (c) and (d) show disagreement between Random Forest and Extra Trees/PointNet++, with the latter two aligning with the expert evaluation.
    }
    \label{tab:model-vs-expert}
    \begin{tabularx}{\linewidth}{l*4{>{\centering\arraybackslash}X}}
    \toprule
    & 
    \includegraphics[width=\linewidth]{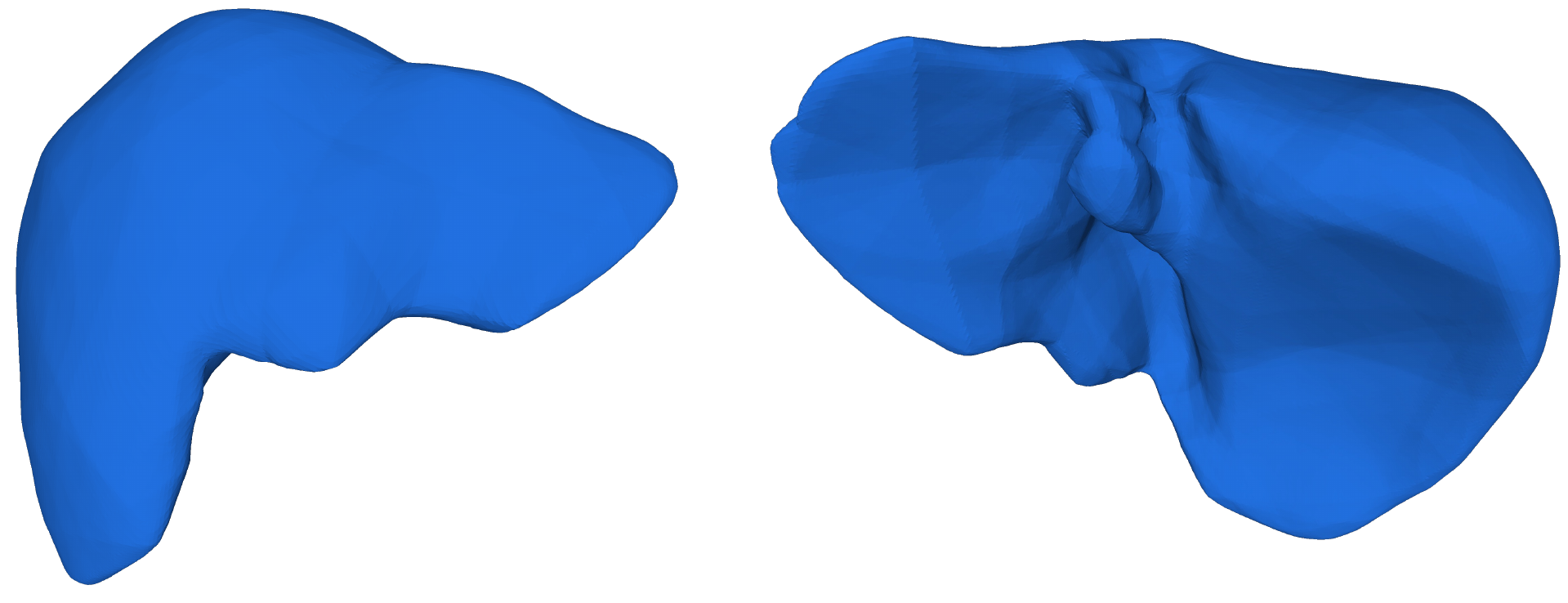} (a) &
    \includegraphics[width=\linewidth]{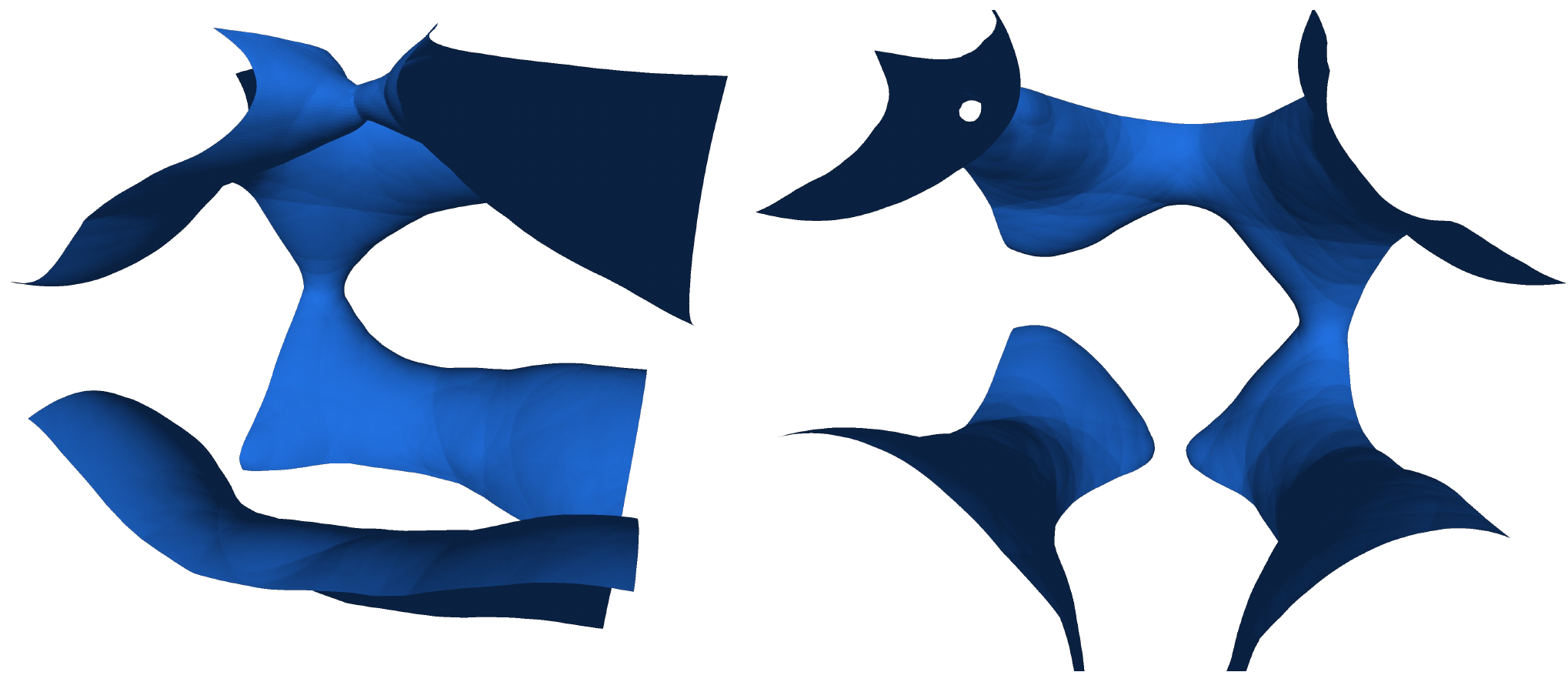} (b) &
    \includegraphics[width=\linewidth]{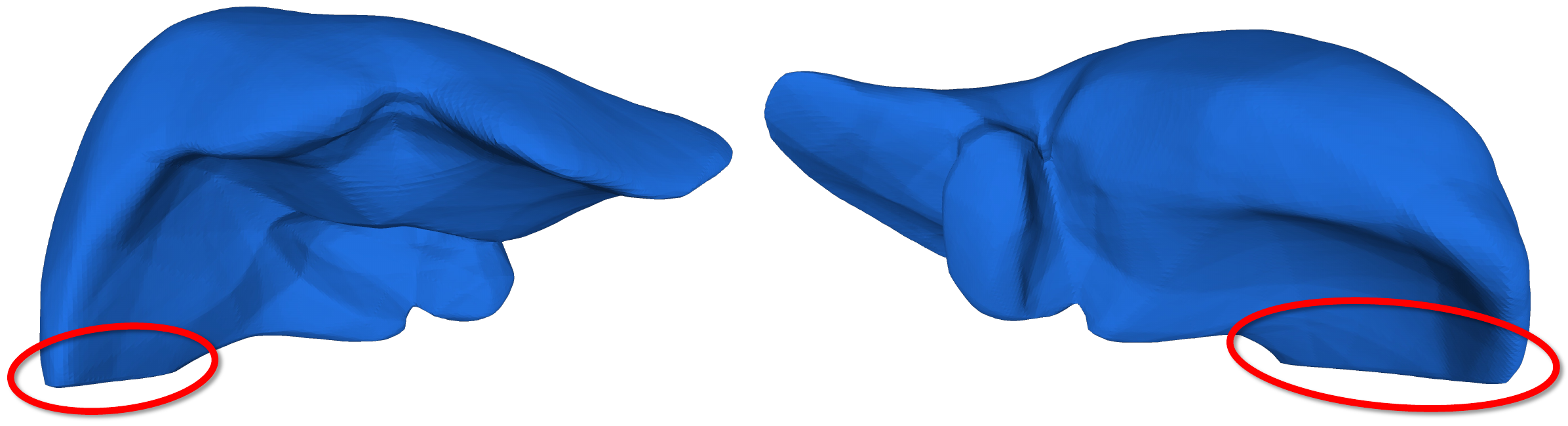} (c) &
    \includegraphics[width=\linewidth]{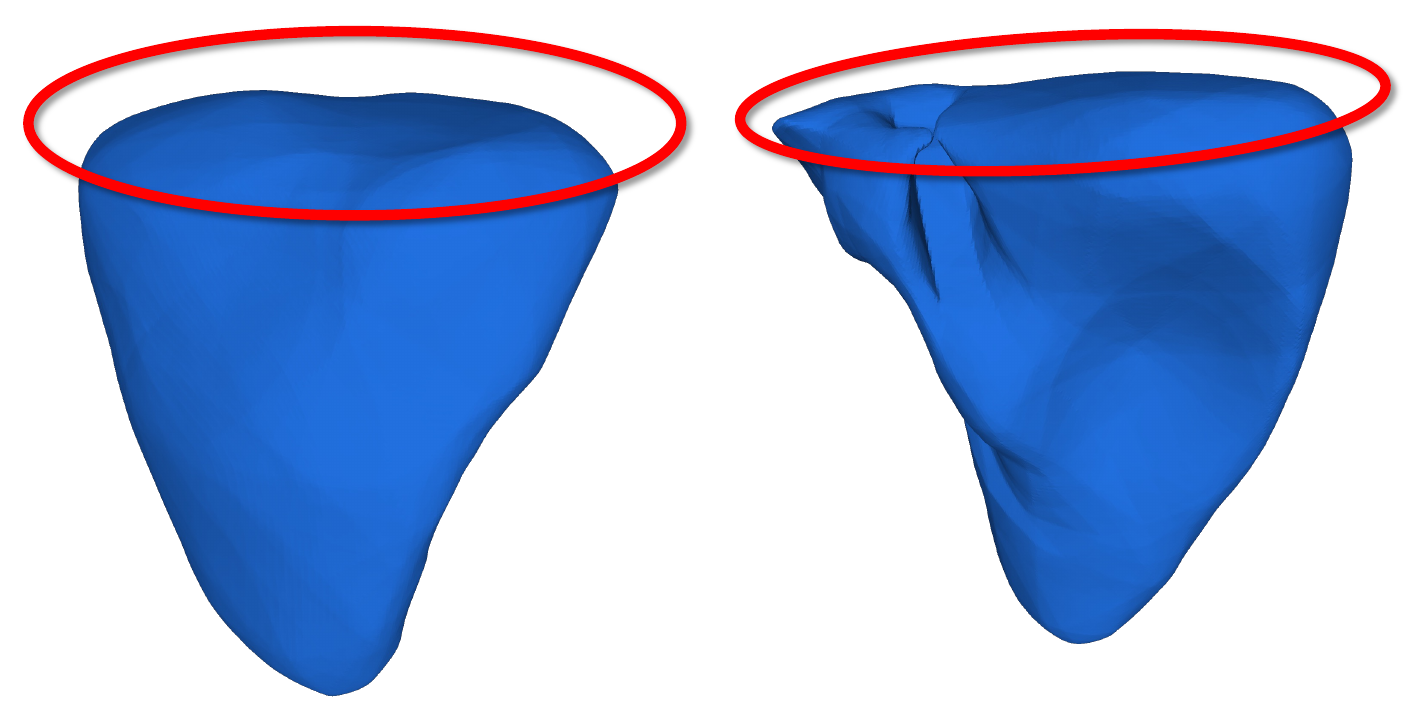} (d) \\ 
    \midrule
    \textbf{Random Forest} & \textcolor{green_good}{Good} & \textcolor{red_bad}{Bad} & \textcolor{red_bad}{Bad}  & \textcolor{red_bad}{Bad} \\
    \textbf{Extra Trees}   & \textcolor{green_good}{Good} & \textcolor{red_bad}{Bad} & \textcolor{red_bad}{Bad}  & \textcolor{green_good}{Good} \\ 
    \textbf{PointNet++}    & \textcolor{green_good}{Good} & \textcolor{red_bad}{Bad} & \textcolor{green_good}{Good} & \textcolor{green_good}{Good} \\ 
    \textbf{Expert}        & \textcolor{green_good}{Good} & \textcolor{red_bad}{Bad} & \textcolor{green_good}{Good} & \textcolor{green_good}{Good} \\ 
    \bottomrule
    \end{tabularx}
\end{table}

To better understand the decision-making process of different classifiers, we present qualitative examples in Table~\ref{tab:model-vs-expert}, showing four representative generated liver objects along with predictions obtained from Random Forest, Extra Trees, and PointNet++, as well as the expert assessments.

Table~\ref{tab:model-vs-expert}(a) represents a case where all methods achieve consensus, correctly identifying a well-formed liver shape as `Good'. 
Similarly, Table~\ref{tab:model-vs-expert}(b) shows universal agreement on a clearly problematic liver with severe fragmentation, classified as `Bad' by all approaches.

More interesting are the disagreement cases. 
In Table~\ref{tab:model-vs-expert}(c), Random Forest and Extra Trees classify the liver as `Bad', likely due to irregular inferior boundary characteristics (highlighted by red circles), while PointNet++ and the expert evaluate it as `Good' based on overall shape acceptability.
This demonstrates that geometric feature-based methods may be more sensitive to specific boundary irregularities.

Table~\ref{tab:model-vs-expert}(d) presents another disagreement case where only Random Forest predicts `Bad', while Extra Trees, PointNet++, and the expert all classify it as `Good'.
This suggests that the 14-dimensional geometric features captured by Random Forest may detect subtle shape anomalies that are not immediately apparent to other methods or deemed clinically acceptable by the expert.

These qualitative examples illustrate the complementary nature of different approaches: geometric feature-based methods (Random Forest, Extra Trees) tend to flag specific structural irregularities, while PointNet++ demonstrates strong alignment with expert clinical judgment. 
The disagreements between methods provide valuable insights into the different aspects of liver shape quality that each approach prioritizes, suggesting potential benefits of combining multiple evaluation perspectives in clinical practice.

\subsection{SHAP-based Feature Importance Analysis}
\label{sec:SHAP}

\begin{figure}[tb]
    \centering
    \includegraphics[width=0.9\textwidth]{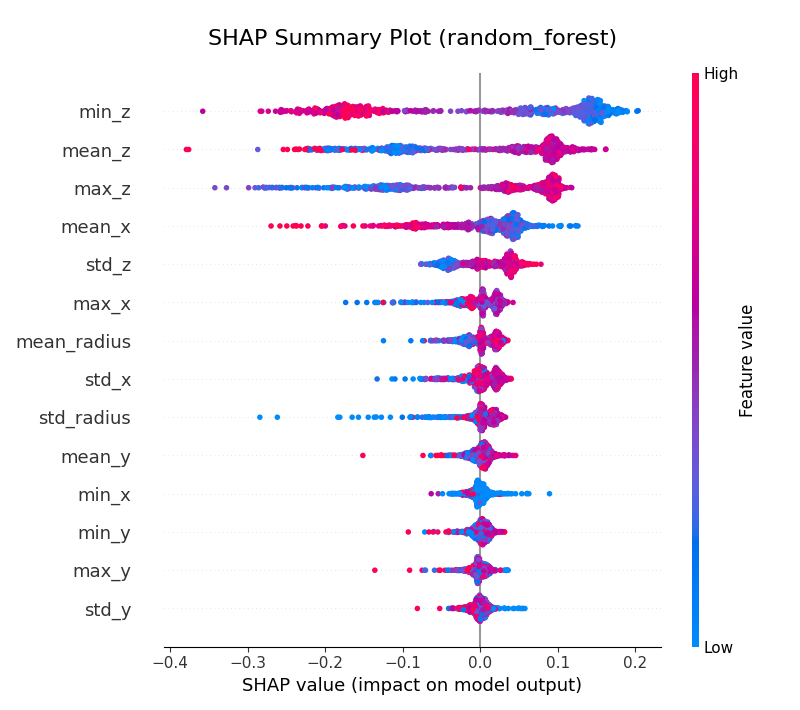}
    \caption{SHAP summary plot showing feature importance and impact on model predictions.}
    \label{fig:shap_summary}
\end{figure}

To identify the geometric features that contribute most to classification decisions, we conducted SHAP (SHapley Additive exPlanations)~\cite{lundberg2017unified} analysis on the Random Forest classifier, which achieved the best predictive performance among classical ML methods. 
Figure~\ref{fig:shap_summary} presents the SHAP summary plot showing both feature importance and the directional impact of each feature on model predictions.

Each feature dimension name is positioned higher according to its importance level. 
Each point represents one liver object, with point colors indicating feature values: red for high values and blue for low values. 
The x-axis represents SHAP values, where points in the positive region of the x-axis increase the likelihood of being predicted as `Good', while points in the negative region increase the likelihood of being predicted as `Bad'.

For example, min\_z, which has the most decisive influence on predictions, shows red points predominantly located in the negative region of the x-axis, while blue points are mostly positioned in the positive region. 
This can be interpreted as higher min\_z values increasing the likelihood of `Bad' predictions, while lower min\_z values increase the likelihood of `Good' predictions.

From an anatomical perspective, the z-axis in CT scans typically represents the cranio-caudal direction (head-to-foot direction), where lower z values indicate the inferior (caudal) direction of the body. 
The tendency for higher min\_z values to be classified as `Bad' may indicate liver objects that are positioned more superiorly (cranially) than anatomically expected, or cases where the inferior boundary of the liver has been inappropriately truncated. 
This is presumably related to errors that can occur during CT image segmentation, incomplete liver boundary extraction, or failure to distinguish boundaries with adjacent organs. 
In particular, the inferior aspect of the liver is adjacent to other organs such as the gallbladder, duodenum, and right kidney~\cite{netter2019atlas}, making it a challenging region for accurate segmentation.

This interpretation is further supported by our qualitative analysis shown in Table~\ref{tab:model-vs-expert}(c), where the Random Forest classifier identifies a liver object as `Bad' due to irregular inferior boundary characteristics (highlighted in red circle), while PointNet++ and expert evaluation classify it as `Good' based on overall shape acceptability. 
This demonstrates that the geometric feature-based approach captures specific boundary-related issues that align with the min\_z feature importance, providing complementary insights to expert evaluation for quality assessment of generated liver shapes.

\section{Discussion and Conclusions}

\subsection{Key Findings and Clinical Implications}

Our results demonstrate several important findings. Random Forest and Extra Trees achieved 90.91\% accuracy in classifying liver point cloud quality, showing that effective quality assessment is possible using only simple 14-dimensional geometric features.

Particularly noteworthy is that PointNet++ achieved perfect agreement with expert evaluation ($\kappa = 1.00$) on the generated dataset, despite showing relatively lower predictive performance (86.7\%) on the test set.
This suggests that PointNet++ possesses capabilities similar to expert clinical judgment in evaluating the quality of newly generated livers based on essential liver features learned from training data.

As mentioned earlier in Section~\ref{sec:SHAP},
the importance of the min\_z feature revealed through SHAP analysis has medical significance. 
Considering that the z-axis represents the cranio-caudal direction in CT scans, the association between higher min\_z values and bad predictions indicates cases where the inferior boundary of the liver has been inappropriately truncated or is abnormally positioned more superiorly than anatomically expected. 
This is related to common segmentation errors that can occur in the inferior aspect of the liver, where boundary distinction with adjacent organs such as the gallbladder, duodenum, and right kidney is challenging.

Such ML-based quality assessment can be utilized in clinical workflows as follows: (1) rapid primary screening of large volumes of generated liver objects, (2) automatic filtering of objects with obvious quality issues before expert review, and (3) providing objective quality criteria for medical staff training.

\subsection{Complementary Nature of ML and Expert Evaluation}

Qualitative analysis reveals the complementary characteristics of each method. While clear cases where all methods agree (Table~\ref{tab:model-vs-expert}a, \ref{tab:model-vs-expert}b) demonstrate reliability, disagreement cases reveal the unique strengths of each approach.

In Table~\ref{tab:model-vs-expert}(c), Random Forest and Extra Trees classified the liver as `Bad' based on irregular inferior boundary characteristics, while PointNet++ and the expert evaluated it as `Good' based on overall shape acceptability.
This shows that geometric feature-based methods can sensitively detect specific structural abnormalities, while deep learning methods and experts place greater emphasis on overall clinical utility.

ML classifiers are particularly useful in the following situations: (1) reducing evaluation bias through the consistent application of different criteria, (2) continuous 24-hour quality monitoring, and (3) providing objective and reproducible quality metrics.
Conversely, expert evaluation remains essential for complex anatomical variations, subtle pathological features, and final quality assurance for clinical application.

\subsection{Interpretability vs Performance Trade-off}

This study demonstrates an important trade-off between interpretability and performance in medical AI. 
Random Forest achieved the highest test performance (90.91\%) and enables clear interpretation through SHAP analysis, while PointNet++ showed perfect expert agreement on generated data despite relatively lower test performance (86.71\%).

The importance of interpretability in the medical domain stems from several factors:

\textbf{Regulatory Requirements}: Regulatory agencies such as the FDA are requiring explainability in medical AI workflows. Transparency in decision-making processes is essential for automated quality assessment approaches to gain clinical approval.

\textbf{Clinical Trust}: For medical staff to trust and appropriately utilize predictive approaches, they must understand the reasoning behind the predictions made. The feature-wise contributions provided by SHAP analysis enable medical staff to verify predictions and interpret them within clinical context.

\textbf{Safety and Error Analysis}: Since incorrect classifications can affect patient care, transparency is crucial for identifying causes when predictions are wrong and for implementing improvements.

In actual clinical applications, a hybrid strategy utilizing the advantages of both approaches may be optimal: using PointNet++ as a primary screening tool and employing Random Forest, along with SHAP analysis, to identify specific quality issues.

\subsection{Limitations and Future Research Directions}

This study has several important limitations, which represent potential directions for future research.

\begin{itemize}
    \item \textbf{Subjectivity of Single Expert Evaluation}: The current study relies on evaluation by a single expert, introducing subjectivity issues. Liver shape quality assessment inherently includes subjective elements, and opinions may differ even among experts. Future research requires multi-expert evaluation and inter-rater reliability analysis.
    
    \item \textbf{Limited Feature Set}: The 14 geometric features currently used are basic and may miss complex anatomical details. Future research should include advanced (engineered) features reflecting liver-specific morphological characteristics (liver lobe segmentation, surface curvature, topological properties).
    
    \item \textbf{Small Generated Dataset}: The limited size of the generated dataset (63 objects) results in insufficient statistical power. 
    Validation on larger-scale datasets generated from various generative models is needed.
    
    \item \textbf{Binary Classification Limitations}: The current `Good/Bad' classification oversimplifies the nuanced grading required in clinical practice.\\ 
    Intermediate-level assessments such as `Usable but requires editing' or \\`Partially useful' may be more clinically valuable.
    
    \item \textbf{Generalizability}: Generalizability across different medical institutions, \\
    scanning protocols, and patient populations has not been validated.
\end{itemize}

\noindent{Further directions for future research are as follows:}

\begin{itemize}
    \item extension to multi-organ studies (e.g., kidney, heart, lungs);
    \item integration of quality metrics into conditional generative models;
    \item construction of lightweight models for real-time quality assessment;
    \item design of efficient expert annotation strategies through active learning; and
    \item exploration of data augmentation techniques for training dataset expansion.
\end{itemize}

\subsection{Conclusions}

This study presented an interpretable approach for evaluating the quality of generated liver point clouds using classical ML methods and PointNet. Random Forest and Extra Trees achieved over 90\% accuracy, while PointNet++ demonstrated perfect agreement with expert assessments. Through SHAP analysis, we were able to interpret the medical significance of geometric features, providing clinically meaningful insights.

ML-based shape classifiers provide not only interpretable feedback but also complementary insights compared to expert evaluation. This study suggests that ML classifiers can serve as lightweight, task-relevant quality metrics in 3D organ shape generation, supporting more transparent and clinically aligned evaluation protocols in medical shape modeling.

Future extension studies with larger datasets, multi-expert evaluation, and advanced anatomical features are expected to further enhance the clinical utility of the presented approach.

%
%
%
\bibliographystyle{splncs04}
\bibliography{ref}
\end{document}